# Learning Item Trees for Probabilistic Modelling of Implicit Feedback

Andriy Mnih

amnih@gatsby.ucl.ac.uk

Gatsby Computational Neuroscience Unit
University College London

Yee Whye Teh

ywteh@gatsby.ucl.ac.uk

Gatsby Computational Neuroscience Unit
University College London

September 23, 2018

**Abstract**

User preferences for items can be inferred from either explicit feedback, such as item ratings, or implicit feedback, such as rental histories. Research in collaborative filtering has concentrated on explicit feedback, resulting in the development of accurate and scalable models. However, since explicit feedback is often difficult to collect it is important to develop effective models that take advantage of the more widely available implicit feedback. We introduce a probabilistic approach to collaborative filtering with implicit feedback based on modelling the user's item selection process. In the interests of scalability, we restrict our attention to tree-structured distributions over items and develop a principled and efficient algorithm for learning item trees from data. We also identify a problem with a widely used protocol for evaluating implicit feedback models and propose a way of addressing it using a small quantity of explicit feedback data.

## 1   Introduction

The rapidly growing number of products available online makes it increasingly difficult for users to choose the ones worth their attention. Recommender systems assist users in making these choices by ranking the products based on inferred user preferences. Collaborative filtering [5] has become the approach of choice for building recommender systems due to its ability to learn complex preference patterns from large collections of user preference data. Most collaborative filtering research deals with inferring user preferences from explicit feedback, for example ratings users gave to items. As a result, several effective methods have been developed for this version of the problem. Matrix factorization based models [4, 11, 12] have emerged as the most popular of these methods due to their simplicity and superior predictive performance. These models are also highly scalable because their training algorithms take advantage of the sparsity of the rating matrix, resulting in training times that are linear in the number of observed ratings.

However, since explicit feedback is often difficult to collect it is essential to develop effective models that take advantage of the more abundant implicit feedback, such as logs of user purchases, rentals, or even clicks. The difficulty of modelling implicit feedback comes from the fact that it contains only positive examples, since users explicitly express their interest (by selecting items) but not their disinterest. Note that not selecting a particular item is not necessarily an expression of disinterest, because it might also be due to the obscurity of the item, lack of time, or other reasons.

Just like their explicit feedback counterparts, the most successful implicit feedback collaborative filtering (IFCF) methods are based on matrix factorization [3, 8, 9]. However, instead of a highly sparse rating matrix, they approximate a dense binary matrix, where each entry indicates whether or not a particular user selected a particular item. We will collectively refer to such methods as Binary Matrix Factorization (BMF). Such approaches amount to treating unobserved user/item pairs as fake negative examples which can dominate the much less numerous positive examples, so the contribution to the objective function from the zero entries is typically downweighted. Since the matrix being approximated is no longer sparse, models of this type are typically trained using batch methods based on alternating least squares. As a result, the training time is cubic in the number of latent factors, which makes these models less scalable than their explicit feedback counterparts.

Recently [10] introduced a new method, called Bayesian Personalized Ranking (BPR), for modelling implicit feedback that is based on more realistic assumptions than BMF. Instead of assuming that users like the selected items and dislike the unselected ones, it assumes that users simply prefer the selected items. The model is presented with selected/unselected item pairs and is trained to rank the selected items above the unselected ones. Since the number of such pairs is typically very large, the unselected items are simply sampled at random.

In this paper we develop a new method that explicitly models the user item selection process using a generative model. Our approach is probabilistic and, unlike other approaches to modelling implicit feedback, can be used to generate new item lists. Like BPR it assumes that selected items are more interesting than the unselected ones. Unlike BPR, however, it represents the appeal of items to a user using a probability distribution, producing a complete ordering of items by probability value. In order to learn distributions over large numbers of items efficiently, we restrict our attention to tree-structured distributions. Since the accuracy of the resulting models depends heavily on the choice of the tree structure, we develop an algorithm for learning trees from data that takes into account the structure of the model the tree will be used with.

We then turn our attention to the task of evaluating implicit feedback models and point out a problem with a widely used evaluation protocol, which stems from the assumption that all items not selected by a user are not relevant. Our proposed solution involves using a small quantity of explicit feedback to reliably identify the not relevant items.

This paper makes three contributions: it proposes a new probabilistic model for implicit feedback, develops a principled model-based algorithm for learning trees over items, and suggests a way to address a problem with the standard evaluation protocol for implicit feedback models.

## 2 Modelling item selection

We propose a new approach to collaborative filtering with implicit feedback based on modelling the item selection process performed by each user. The identities of the items selected by a user are modelled as independent samples from a user-specific distribution over all available items. The probability of an item under this distribution reflects the user's interest in it. Training our model amounts to performing multinomial density estimation for each user. As a result, the model naturally learns from the observed user / item pairs without explicitly considering the unobserved pairs.

To make the modelling task more manageable we make two simplifying assumptions. First, we assume that user preferences do not change with time and model all items chosen by a user as independent samples from a fixed user-specific distribution. Second, to keep the model as simple as possible we assume that items are sampled with replacement. We believe that sampling with replacement is a reasonable approximation to sampling without replacement in this case because the space of items is large while the number of items selected by a user is relatively small, from which it follows that the probability of an item being selected more than once by the same user will be small. These simplifications allow us to model the identities of the items selected by a user as IID samples.

We now outline a simple implementation of the proposed idea which, though impractical for large datasets, will serve as a basis for developing a more scalable model. As is typical for matrix factorization methods in collaborative filtering, we represent users and items with real-valued vectors of latent factors. The factor vectors for user $u$ and item $i$ will be denoted by $U_u$ and $V_i$ respectively. Intuitively, $U_u$ captures the preferences of user $u$, while $V_i$ encodes the properties of item $i$. Both user and item factor vectors are unobserved and so have to be learned from the observed user / item pairs. The dot product between $U_u$ and $V_i$ quantifies the preference of the user for the item. We define the probability of user $u$ choosing item $i$ as

$$P(i|u) = \frac{\exp(U_u^\top V_i + c_i)}{\sum_k \exp(U_u^\top V_k + c_k)}, \tag{1}$$

where $c_i$ is the bias parameter that captures the overall popularity of item $i$ and index $k$ ranges over all items in the inventory. The model can be trained using stochastic gradient ascent [2] on the log-likelihood by iterating through the user / item pairs in the training set, updating $U_u$, $V_i$, and $c_i$ based on the gradient of $\log P(i|u)$.

The main weakness of the model is that its training time is linear in the inventory size because computing the gradient of the log-probability of a single item requires explicitly considering all available items. Though linear time complexity might not seem prohibitive, it greatly limits the applicability of the model since collaborative filtering tasks with tens or even hundreds of thousands of items are now common.

# 3  Hierarchical item selection model

The linear time complexity of the gradient computation is a consequence of normalization over the entire inventory in Eq. 1, which is required because the space of items is unstructured. We can speed up normalization, and thus learning, exponentially by assuming that the space of items has a known tree structure.

We start by supposing that we are given a $K$-ary tree with items at the leaves with exactly one item per leaf. For simplicity, we will assume that each item is located at exactly one leaf. Such a tree is uniquely determined by specifying for each item the path from the root to the leaf containing the item. Any such path can be represented by the sequence of nodes $n = n_1, ..., n_L$ it visits. Since all paths we consider start at the root, we leave the root out, starting the sequence instead with the second node in the path, which is a child of the root. Since we only deal with paths where node $n_{j+1}$ is a child of node $n_j$ for all $j$, we can also uniquely determine the identity of node $n_{i+1}$ by specifying which of the children of the previous node it is. This leads to an alternative encoding $d = d_1, ..., d_L$ of $n$ with the property that node $n_{j+1}$ is child number $d_{j+1}$ of node $n_j$. For example, $d^i = 4, 1$ indicates that item $i$ is reached by starting at the root, then visiting its fourth child, and then visiting the first child of that child. We will refer to $d^i$ as the *code* for item $i$ and to the elements of $d^i$ as *digits*[1]. Since codes for different items can have different length, we will denote the length of the code for item $i$ by $L_i$. In the following, $n$ and $d$ (or $n^i$ and $d^i$) will always refer to the two representations of the same code.

By making the choice of the next node stochastic, we can induce a distribution over the leaf nodes in the tree and thus over items. To allow each user to have a different distribution over items we make the probability of choosing each child of a function of the user's factor vector. The probability will also depend on the child node's factor vector and bias the same way the probability of choosing an item in Eq. 1 depends on the item's factor vector and bias.

Before defining the distribution over children of a node we first need to introduce some notation. Let $C(n_j)$ be the set of children of node $n_j$ and $C(n_j, k)$ be its $k^{th}$ child. Let $Q_{n_j}$ and $b_{n_j}$ be the node's factor vector and bias respectively. Then for user $u$ the probability of choosing the $k^{th}$ child of node $n_j$ is given by

$$P(d_{j+1} = k | n_j, u) = \frac{\exp\left(U_u^\top Q_{C(n_j, k)} + b_{C(n_j, k)}\right)}{\sum_{m \in C(n_j)} \exp\left(U_u^\top Q_m + b_m\right)}, \quad (2)$$

where $b_{n_j}$ is the bias of node $n_j$ that captures its base rate. We define $n_0$ to be the root node so that the above probability is defined for $j = 0$.

The probability of selecting item $i$ is then given by the product of the probabilities of making the sequence of decisions that leads from the root to the leaf containing $i$:

$$P(i|u) = \prod_{j=1}^{L_i} P(d_j^i | n_{j-1}^i, u). \quad (3)$$

We will call the model defined by Eq. 3 the Collaborative Item Selection (CIS) model. Given a tree over items, the CIS model can be trained using stochastic gradient ascent in log-likelihood, updating parameters after each user / item pair. In fact, training the model reduces to training a collection of multinomial logistic regression (MLR) models, one per non-leaf node, with the inputs to the models learned jointly with model parameters. From this point of view, the parameters of the MLR model associated with a node are the factor vectors and biases of the node's children, while the model's inputs are the user factor vectors.

While the model can be based on any tree over items, the choice of the tree affects the model's efficiency and its ability to generalize. Since computing the probability of a single item takes time linear in the item's depth in the tree, we want to avoid trees that are too unbalanced. To produce a model that generalizes well we also want to avoid trees with difficult classification problems at the internal nodes [1], which correspond to hard-to-predict item codes. Thus a random tree, even if it is a balanced one, is unlikely to be a good choice because it requires classifying items into $K$ random classes, a task with very little structure when the number of items in each class is large.

One way to produce a tree that results in relatively easy classification problems is to assign similar items to the same class, which is the approach of [6] and [13]. However, the similarity metrics used by these methods are not model-based in the sense that they are not derived from the classifier models that will be used at the tree nodes. We would like to develop a scalable model-based algorithm for learning trees that correspond to item codes that are easy to predict using Eq. 2.

---

[1] These are base-$(K + 1)$ digits since we are using them to number children in a $K$-ary tree, counting from 1.

# 4 Related work

The use of tree-structured label spaces to reduce the normalization cost has originated in statistical language modelling, where it was used to accelerate neural language models [7].

The task of learning trees for efficient probabilistic multiclass classification has received surprisingly little attention. The two algorithms most closely related to the one proposed in this paper are [1] and [6]. [1] proposed a fully online algorithm for multinomial density estimation that constructs a binary label tree by inserting the previously unseen labels whenever they are encountered. The location for a new label is found proceeding from the root to a leaf making the left child / right child decisions based on their probability under the model and a tree balancing penalty. This is the only tree learning algorithm we are aware of that takes into account the probabilistic model the tree is used with. Unfortunately, this approach is very optimistic because it decides on the location for a new label in the tree based on a single training case and never revisits that decision.

The algorithm in [6] was developed for learning trees over words for use in probabilistic language models. It constructs such trees by performing top-down hierarchical clustering of words, which are represented by real-valued vectors. The word representations are learned through bootstrapping by training a language model based on a random tree. This algorithm, unlike the one we propose in Section 5, does not take into consideration the model the tree is constructed for.

Most work on tree-based multiclass classification deals with non-probabilistic models and does not apply to the problem we are concerned with in this paper. Of these approaches our algorithm is most similar to the one in [13], which looks for a tree structure that avoids requiring to discriminate between easily confused items as much as possible. The main weakness of that approach is the need for training a flat classifier to produce the confusion matrix needed by the algorithm. As a result, it is unlikely to scale to large datasets containing tens of thousands of classes.

# 5 Learning item codes: the model-based approach

## 5.1 Overview

In this section we develop a scalable algorithm for learning trees that takes into account the parametric form of the model the tree will be used with. At the highest level our approach can be seen as top-down model-based hierarchical clustering of items. We chose top-down clustering over bottom-up clustering because it is the more scalable option. Since finding the best tree is intractable, we take a greedy approach that constructs the tree one level at a time, learning the $l^{th}$ digit of all item codes before fixing it and advancing to the next digit. Because our approach is model-based, it learns model parameters, i.e. node biases and factor vectors, jointly with the item codes. As a result, at every point during its execution it specifies a complete probabilistic model of the data, which gets progressively more expressive with each tree level learned. As a result, that we can monitor progress of the algorithm by evaluating the predictions made after learning each layer.

For simplicity, our tree learning algorithm assumes that user factor vectors are known and fixed. Since these vectors are actually unknown, we learn them by first training a CIS model based on a random balanced tree. We then extract the user vectors learned by the model and use them to learn a better tree from the data. Finally, we train a CIS model based on the learned tree. This three-stage approach is similar to the one used in [6] to learn trees over words. However, because our tree learning algorithm is model-based, we already have a complete probabilistic model at its termination, so we only need to finetune its parameters instead of learning them from scratch. Finetuning is necessary because the parameters learned while building the tree are based on the fixed user factor vectors from the random-tree-based model.

## 5.2 Learning a level of a tree

We now describe how to learn a level of the tree. Suppose we have learned the first $l-1$ digits of each item code and would like to learn the $l$'s digit. Let $U_i$ be the set of users who rated item $i$ in the training set. The contribution made by item $i$ to the log-likelihood is then given by

$$L_i = \log \prod_{u \in U_i} P(i|u) = \sum_{u \in U_i} \log \prod_j P(d_j^i | n_{j-1}^i, u) = \sum_{u \in U_i} \sum_j \log P(d_j^i | n_{j-1}^i, u). \tag{4}$$

The log-likelihood contribution due to a single observation can be expressed as

$$\sum_j \log P(d_j^i | n_{j-1}^i, u) = \sum_{j=1}^{l-1} \log P(d_j^i | n_{j-1}^i, u) + \log P(d_l^i | n_{l-1}^i, u) + \sum_{j=l+1}^{L_i} \log P(d_j^i | n_{j-1}^i, u). \quad (5)$$

The first term on the RHS depends only on the parameters and code digits that have already been learned, so it can be left out of the objective function. The third term is the log-probability of item $i$ under the subtree rooted in node $n_l^i$, which depends on the structure and parameters of that subtree, which we have not learned yet. To emphasize the fact that this term is based on a user-dependent distribution over items under node $n_l^i$ we will refer to it as $\log P(i | n_l^i, u)$. Note that $n_l^i$ depends on $d_l^i$, as it is the $(d_l^i)^{th}$ child of $n_{l-1}^i$.

The overall objective function for learning level $l$ is obtained by adding up the contributions of all items, leaving out the terms that do not depend on the quantities to be learned:

$$L^l = \sum_i \sum_{u \in U_i} \log P(d_l^i | n_{l-1}^i, u) + \sum_i \sum_{u \in U_i} \log P(i | n_l^i, u). \quad (6)$$

The most direct approach to learning the codes would be to alternate between updating the $l^{th}$ digit in the codes and the factor vectors (and biases) of the $l^{th}$ level nodes'. Since jointly optimizing over the $l^{th}$ digit in all item codes is infeasible, we have to resort to incremental updates, maximizing $L^l$ over the $l^{th}$ digit in one item code at a time. Unfortunately, even this operation is intractable because evaluating each value of $d_l^i$ requires knowing the optimal contribution from the still-to-be-learned levels of the tree, which is the second term in Eq. 6. In other words, to find the optimal $d_l^i$ we need to compute

$$d_l^i = \arg\max_d \left( \sum_{u \in U_i} \log P(d | n_{l-1}^i, u) + F(d, n_{l-1}^i) \right), \quad (7)$$

where we left out the terms that do not depend on $d_l^i$. The optimal contribution $F(d_l^i, n_{l-1}^i)$ from the future levels is defined as

$$F(d_l^i, n_{l-1}^i) = \max_\Theta \sum_{k \in I(n_{l-1}^i)} \sum_{u \in U_k} \log P(k | n_l^k, u), \quad (8)$$

where $I(n_{l-1}^i)$ is the set of items that are assigned to node $n_{l-1}^i$, and $\Theta$ is the set of node factor vectors, biases, and tree structures that parameterize the set of distributions $\{P(k|n_l^k, u) | k \in I(n_{l-1}^i)\}$.

### 5.3 Approximating the future

The value of $F(d_l^i, n_{l-1}^i)$ quantifies the difficulty of discriminating between items assigned to node $n_{l-1}^i$ using the best tree structure and parameter setting possible given that item $i$ is assigned to the $(d_l^i)^{th}$ child of that node. Since $F(d_l^i, n_{l-1}^i)$ in Eq. 8 rules out degenerate solutions where all items below a node are assigned to the same child of it, leaving $F(d_l^i, n_{l-1}^i)$ out to make the optimization problem easier is not an option.

We address the intractability of Eq. 7 while avoiding the degenerate solutions by approximating the tree-structured distributions $P(k|n_l^k, u)$ by simpler distributions that make it much easier to evaluate $F(d_l^i, n_{l-1}^i)$ for each candidate value for $d_l^i$. Since computing $F(d_l^i, n_{l-1}^i)$ requires maximizing over the free parameters of $P(k|n_l^k, u)$, choosing a parameterization of $P(k|n_l^k, u)$ that makes this maximization easy can greatly speed up this computation. We propose replacing the tree-structured $P(k|n_l^k, u)$ by a flat user-independent distribution $P(k|n_l^k)$. The main advantage of this parameterization is that the optimal $P(k|n_l^k)$ can be computed by counting the number of times each item assigned to node $n_l^k$ occurs in the training data and normalizing. In other words, when $P(k|n_l^k)$ is used in Eq. 8 the maximum is achieved at

$$P(i|n_l^k) = \begin{cases} \frac{N_i}{\sum_{m \in I(n_l^k)} N_m} & \text{if } i \in I(n_l^k) \\ 0 & \text{otherwise} \end{cases} \quad (9)$$

where $N_i$ is the number of times item $i$ occurs in the training set. The corresponding value for $F(d_l^i, n_{l-1}^i)$ is given by

$$F(d_l^i, n_{l-1}^i) = \sum_{k \in I(n_{l-1}^i)} N_k \log \frac{N_k}{\sum_{m \in I(n_l^k)} N_m}. \quad (10)$$

To show that $F(d_l^i, n_{l-1}^i)$ can be computed in constant time we start by observing that the sum over items under node $n_{l-1}^i$ can be written in terms of sums over items under each of its child nodes:

$$F(d_l^i, n_{l-1}^i) = \sum_{c \in C(n_{l-1}^i)} \sum_{k \in I(c)} N_k \log \frac{N_k}{\sum_{m \in I(c)} N_m}$$

$$= \sum_{c \in C(n_{l-1}^i)} \sum_{k \in I(c)} N_k \log N_k - \sum_{c \in C(n_{l-1}^i)} Z_c \log Z_c. \quad (11)$$

with $Z_c = \sum_{k \in I(c)} N_k$. Since adding a constant to $F(d_l^i, n_{l-1}^i)$ has no effect on the solution of Eq. 7 and the first term in the equation does not depend on $d_l^i$, we can drop it to get

$$\tilde{F}(d_l^i, n_{l-1}^i) = \sum_{c \in C(n_{l-1}^i)} Z_c \log Z_c. \quad (12)$$

To compute $\tilde{F}(d_l^i, n_{l-1}^i)$ efficiently, we store $Z_c$'s and the old $\tilde{F}(d_l^i, n_{l-1}^i)$ value, updating them whenever an item is assigned to a different node. Such updates can be performed in constant time.

We now show that the first term in Eq. 7, corresponding to the contribution of the $l^{th}$ code digit for item $i$, can be computed efficiently. Plugging in the definition of $P(d_{j+1} = k|n_j, u)$ from Eq. 2 we get

$$\sum_{u \in U_i} \log P(d|n_{l-1}^i, u) = \sum_{u \in U_i} \left( U_u^\top Q_{C(n_j^i, d)} + b_{C(n_j^i, d)} \right) + C$$

$$= \left( \sum_{u \in U_i} U_u \right)^\top Q_{C(n_j^i, d)} + |U_i| b_{C(n_j^i, d)} + C \quad (13)$$

where $C$ is a term that does not depend on $d_l^i$ and so does not have to be considered when maximizing over $d_l^i$. Since we assume that the user factor vectors are known and fixed, we precompute $R_i = \sum_{u \in U_i} U_u$ for each user, which can be seen as creating a surrogate representation for item $i$.

Plugging Eq. 13 into Eq. 7 gives us the equation for updating item digits:

$$d_l^i = \arg\max_d \left( R_i^\top Q_{C(n_j, d)} + |U_i| b_{C(n_j, d)} + \tilde{F}(d, n_{l-1}^i) \right). \quad (14)$$

## 6 Evaluating models of implicit feedback

Establishing sensible evaluation protocols for a machine learning problem is important because by specifying how the performance of a method should be evaluated, they effectively define what "better" performance means and implicitly guide the development of future methods. Given that the problem of implicit feedback collaborative filtering is relatively new it is not surprising that the typical evaluation protocol was adopted from information retrieval. However, we believe that this protocol is much less well suited for collaborative filtering than it is for information retrieval.

Implicit feedback models are typically evaluated using information retrieval metrics such as Mean Average Precision (MAP), which requires knowing which items are relevant and which are not relevant to each user. It is typical to assume that the items the user selected are relevant and all others are not [9]. However, this approach is problematic because it fails to distinguish between the items the user really has no interest in (i.e. the truly not relevant ones) and relevant items the user simply did not rate. While it might be true that, as is commonly argued, that the not relevant items dominate the unobserved relevant ones, the effect of the latter might still be significant for comparing different models. We demonstrate this in the next section. We propose using some explicit feedback information to identify a small number of truly not relevant items for each user and using them in place of items of unknown relevance in the evaluation. Thus the models will be evaluated on their ability to rank the truly relevant items ahead of the truly not relevant ones, which we believe is the fundamental task of collaborative filtering. Though this approach does require using explicit feedback information, only a small quantity of it is necessary, and it is used only for evaluation.

## 7 Experimental results

We evaluated our algorithms on the large version of the MovieLens dataset which contains 10M ratings on a scale from 0 to 5 assigned by 69878 users to 10677 movies. To simulate the implicit feedback setting, where the presence of a user/item pair indicates an expression of interest, we kept only the user/item pairs associated with ratings 4 and above (and discarded the rating values). We then split the resulting 5M pairs into a 4M-pair training set, and a validation and test sets of 500K pairs each. We compared the models based on their ranking performance, as measured by the standard information retrieval metrics: Mean Average Precision (MAP), Precision@k, Recall@k, as well as Expected Percentile Rank (EPR) [3], which is the average normalized rank of relevant items. We used the evaluation approach described in the previous section, which involved having the models rank only the items with known relevance status. We used the rating values to determine relevance, considering items rated below 3 as not relevant and items rated 4 and above as relevant.

Table 1: Test set scores in percent on the MovieLens dataset obtained by treating items with low ratings as not relevant. EPR is Expected Percentile Rank. Higher scores indicate better performance for all metrics except for EPR.

| Model | MAP | EPR | P@1 | P@5 | P@10 | R@1 | R@5 | R@10 |
|---|---|---|---|---|---|---|---|---|
| CIS (Random) | 70.68 | 28.18 | 74.65 | 58.02 | 49.91 | 20.66 | 60.02 | 77.31 |
| CIS (LearnedRI) | 72.50 | 26.97 | 76.64 | 59.29 | 50.64 | 21.51 | 61.24 | 78.22 |
| CIS (LearnedCI) | 72.61 | 26.90 | 76.68 | 59.37 | 50.69 | 21.54 | 61.31 | 78.27 |
| BPR | 72.75 | 26.52 | 75.75 | 59.15 | 50.63 | 21.50 | 61.43 | 78.39 |
| BMF | 70.80 | 28.20 | 75.66 | 58.03 | 49.77 | 20.94 | 60.04 | 77.21 |

Table 2: Test set scores in percent on the MovieLens dataset obtained by treating all unobserved items as not relevant.

| Model | MAP | EPR | P@1 | P@5 | P@10 | R@1 | R@5 | R@10 |
|---|---|---|---|---|---|---|---|---|
| BPR | 12.73 | 2.15 | 14.27 | 11.56 | 9.89 | 3.06 | 11.55 | 18.86 |
| BMF | 16.13 | 3.23 | 22.10 | 16.25 | 12.94 | 4.66 | 15.64 | 23.55 |

We compared our hierarchical item selection model to two state-of-the-art models for implicit feedback: the Bayesian Personalized Ranking model (BPR) and the Binary Matrix Factorization model (BMF). All models used 25-dimensional factor vectors, as we found that higher-dimensional factor vectors resulted in only marginal improvements. We included three CIS models based on different binary trees to highlight the effect of tree construction methods. The methods are as follows: "Random" generates random balanced trees; "LearnedRI" is the method from Section 5 with randomly initialized item-node assignments; "LearnedCI" is the same method with item-node assignments initialized by clustering surrogate item representations $R_i$ from Section 5.3.

Better performance corresponds to lower values of EPR and higher values of the other metrics. Table 1 shows the test scores for the resulting models. CIS (Learned) and BPR emerge as the best performing methods, achieving very similar scores across all metrics. BPR has a slight edge over CIS on MAP and EPR, while CIS performs better on Precision@1. BMF and CIS (Random) are the weakest performers, with considerably worse scores than BPR or CIS (Learned) on all metrics. These results confirm that the quality of the trees used has a strong effect on the performance of CIS models and that using trees learned by the proposed algorithm makes CIS competitive with the best collaborative filtering models. The similar results achieved by CIS (LearnedRI) and CIS (LearnedCI) suggest that that the performance of the resulting model is not particularly sensitive to the initialization scheme of the tree learning algorithm.

To highlight the importance of excluding items of unknown relevance when evaluating implicit feedback models we recomputed the performance metrics with all items not rated by a user treated as not relevant. As the scores in Table 2 show this seemingly minor modification of the evaluation protocol makes BMF appear to outperform BPR by a large margin, which, as Table 1 indicates in not actually the case. In retrospect, these changes in relative performance are not particularly surprising since the training algorithm for BMF treats unobserved items as negative examples, which perfectly matches the assumption the evaluation is based on, namely that unobserved items are not relevant. This is a clear example of a flawed evaluation protocol favouring an unrealistic modelling assumption.

## 8 Discussion

We proposed a model that in addition to being competitive with the best implicit feedback models in terms of predictive accuracy has the advantage of providing calibrated item selection probabilities for each user, which quantify the user's interest in items. These probabilities can be used in combination with a utility function to make sophisticated recommendations, such as cost-sensitive ones.

We believe that evaluation protocols for implicit feedback models deserve more attention than they have received. In this paper we observed that one widely used protocol can produce misleading results due to an unrealistic assumption it makes about item relevance. We proposed using a small quantity of explicit feedback data to directly estimate item relevance in order to avoid having to make that assumption.

Although we introduced our tree learning algorithm in the context of collaborative filtering, it is applicable to several other problems. One such problem is statistical language modelling, where the task is predicting the distribution of the next word in a sentence given its context that consists of several preceding words. While there already exists an algorithm for learning the structure of tree-based language models [6], it constructs trees by clustering word representations, which does not take into account the form of the model that will use these trees. In contrast, our algorithm optimizes the tree structure and model parameters jointly, which can lead to superior model performance.

The proposed algorithm can also be used to learn trees over labels for multinomial regression models. For models with a large number of labels, using a label space with a sensible tree structure can lead to much faster training and improved generalization. Our algorithm can be applied in this setting by noticing the correspondence between labels and items and input vectors and user factor vectors. However, unlike in collaborative filtering where user factor vectors have to be learned, in this case input vectors are observed, which eliminates the need to train a model with a random tree before applying the tree-learning algorithm.

# References

[1] Alina Beygelzimer, John Langford, Yuri Lifshits, Gregory B. Sorkin, and Alexander L. Strehl. Conditional probability tree estimation analysis and algorithms. In *Proceedings of the 25th Conference on Uncertainty in Artificial Intelligence*, 2009.

[2] Léon Bottou. Large-scale machine learning with stochastic gradient descent. In *Proceedings of the 19th International Conference on Computational Statistics (COMPSTAT'2010)*, pages 177–187, 2010.

[3] Yifan Hu, Yehuda Koren, and Chris Volinsky. Collaborative filtering for implicit feedback datasets. In *Proceedings of the 2008 Eighth IEEE International Conference on Data Mining*, pages 263–272, 2008.

[4] Yehuda Koren. Factorization meets the neighborhood: a multifaceted collaborative filtering model. In *Proceeding of the 14th ACM SIGKDD international conference on Knowledge discovery and data mining*, pages 426–434, 2008.

[5] Benjamin Marlin. Collaborative filtering: A machine learning perspective. Master's thesis, University of Toronto, 2004.

[6] Andriy Mnih and Geoffrey Hinton. A scalable hierarchical distributed language model. In *Advances in Neural Information Processing Systems*, volume 21, 2009.

[7] Frederic Morin and Yoshua Bengio. Hierarchical probabilistic neural network language model. In *AISTATS'05*, pages 246–252, 2005.

[8] Rong Pan and Martin Scholz. Mind the gaps: weighting the unknown in large-scale one-class collaborative filtering. In *KDD*, pages 667–676, 2009.

[9] Rong Pan, Yunhong Zhou, Bin Cao, Nathan Nan Liu, Rajan M. Lukose, Martin Scholz, and Qiang Yang. One-class collaborative filtering. In *ICDM*, pages 502–511, 2008.

[10] Steffen Rendle, Christoph Freudenthaler, Zeno Gantner, and Schmidt-Thieme Lars. BPR: Bayesian personalized ranking from implicit feedback. In *UAI '09*, pages 452–461, 2009.

[11] Ruslan Salakhutdinov and Andriy Mnih. Probabilistic matrix factorization. In *Advances in Neural Information Processing Systems*, volume 20, 2008.

[12] Nathan Srebro, Jason D. M. Rennie, and Tommi Jaakkola. Maximum-margin matrix factorization. In *Advances in Neural Information Processing Systems*, 2004.

[13] Jason Weston, Samy Bengio, and David Grangier. Label embedding trees for large multi-class tasks. In *Advances in Neural Information Processing Systems (NIPS)*, 2010.